\documentclass[letterpaper, 10 pt, conference]{ieeeconf}

\IEEEoverridecommandlockouts                              

\usepackage{epsfig} %
\usepackage{amsmath} %
\usepackage{amssymb}  %
\usepackage{times}
\usepackage{graphicx}
\usepackage{color}
\usepackage{stfloats}
\usepackage{gensymb}
\usepackage{subfig}
\usepackage{pifont}
\usepackage{float}

\title{\LARGE \bf
Radar-Camera Sensor Fusion for Joint Object Detection \\and Distance Estimation in 
Autonomous Vehicles
}

\author{Ramin Nabati$^{1}$ and Hairong Qi$^{1}$%
\thanks{$^{1}$Department of Electrical Engineering and Computer Science, The 
University of Tennessee, Knoxville, USA. Email:
{\tt\small mnabati@vols.utk.edu, hqi@utk.edu}
}%
}

\newcommand{\lidar}{LIDAR}

\begin{document}

\maketitle
\thispagestyle{empty}
\pagestyle{empty}
\begin{abstract}
In this paper we present a novel radar-camera sensor fusion framework for accurate 
object detection and distance estimation in autonomous driving scenarios. The proposed 
architecture uses a middle-fusion approach to fuse the radar point clouds and 
RGB images. Our radar object proposal network uses radar point clouds to generate 3D proposals 
from a set of 3D prior boxes. These proposals are mapped to the image and fed 
into a Radar Proposal Refinement (RPR) network for objectness score prediction and 
box refinement. The RPR network utilizes both radar information and image feature 
maps to generate accurate object proposals and distance estimations.

The radar-based proposals are combined with image-based proposals
generated by a modified Region Proposal Network (RPN). The RPN has a distance regression 
layer for estimating distance for every generated proposal. The radar-based and 
image-based proposals are merged and used in the next stage for object classification. 
Experiments on the challenging nuScenes dataset show our method outperforms other
existing radar-camera fusion methods in the 2D object detection task while at the same
time accurately estimates objects' distances.

\end{abstract}

\section{INTRODUCTION}  \label{intro}

Object detection and depth estimation is a crucial part of the perception 
system in autonomous vehicles. Modern self driving cars are usually equipped with 
multiple perception sensors such as cameras, radars and \lidar{}s. Using multiple
sensor modalities provides an opportunity to exploit their complementary properties. 
Nonetheless, the process of multi-modality fusion also makes designing 
the perception system more challenging. Over the past 
few years many sensor fusion methods have been proposed for autonomous
driving applications. Most existing sensor fusion algorithms focus on combining 
RGB images with 3D \lidar{} point clouds \cite{feng2020deep}. \lidar{}s provide 
accurate depth information that could be used for 3D object detection. This is particularly 
useful in autonomous driving applications where having the distance to all detected 
objects is crucial for safe operation.

\begin{figure}[b]
   \centering
   {\setlength{\fboxsep}{0pt}
      \subfloat[]{
         \framebox{%
            \includegraphics[width=0.45\textwidth]{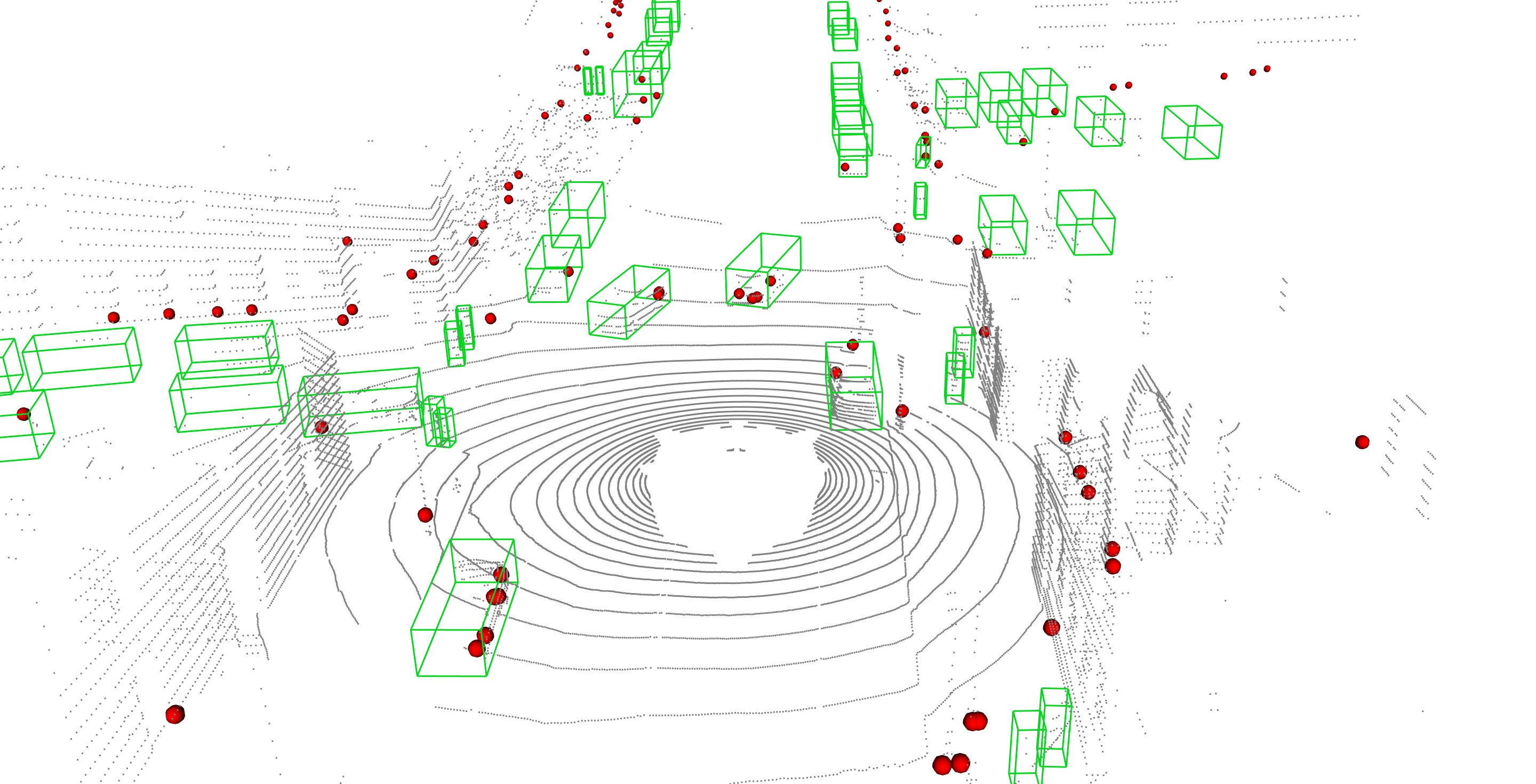}}
            \label{2}
         }
   }
   \caption{Sample data from the NuScenes dataset showing Radar point cloud 
      (red), 3D ground truth boxes (green) and \lidar{} point cloud (grey).}
   \label{fig:NuscSample}
\end{figure}

While \lidar{}s are becoming popular in autonomous vehicles, 
radars have been used in autonomous and also non-autonomous vehicles for many years
as an indispensable depth sensor. Radars operate by 
measuring the reflection of radio waves from objects, and use the Doppler effect 
to estimate objects' velocity. Although radars provide accurate distance and 
velocity information, they are not particularly good at classifying objects. 
This makes the fusion of radar and other sensors such as cameras a very 
interesting topic in autonomous driving applications. A radar-camera fusion system 
can provide valuable depth information for all detected objects in an autonomous 
driving scenario, while at the same time eliminates the need for computationally 
expensive 3D object detection using \lidar{} point clouds.

Due to their unstructured nature, processing depth sensor data is a very challenging 
problem. Additionally, the point cloud obtained by depth sensors are usually sparse with 
very variable point density. In \lidar{} point clouds for example, nearby 
objects have significantly more measurements than far away objects.
This makes the point cloud-based object detection a challenging task.
To overcome this problem, some methods apply 
image-based feature extraction techniques by projecting the point cloud into a 
perspective view \cite{mv3d_2017,AVOD_2018,li2016vehicle}, e.g. the bird's eye 
view (BEV). Other methods \cite{li2016vehicle,lang2018pointpillars,yan2018second} 
partition the point cloud into a regular grid of equally spaced voxels, and then
learn and extract voxel-level features. More recently, Qi \textit{et al.} 
\cite{qi2017pointnet,qi2017pointnet++} proposed PointNet, an end-to-end 
deep neural network for learning point-wise features directly from point clouds 
for segmentation and classification.

Although point cloud feature extraction and classification methods have proven to be
very effective on dense point clouds obtained from \lidar{}s, they are not as effective 
on sparse radar point clouds. For one object, an ideal radar
only reports one point, compared to tens or hundreds of points obtained by a 
\lidar{} for the same object. Additionally, most automotive radars do not provide 
any height information for the detected objects, essentially making the radar 
point clouds a 2-dimensional signal, as opposed to the 3-dimensional point clouds
obtained from a \lidar. 
Another difference between radar and \lidar{} point clouds is the amount of processing 
needed to extract useful features. Automotive radars have built-in functionalities
to extract very useful features for every detection, such as relative 
object speed, detection validity probability and stationary or moving classification
for objects. While one can use these features directly without any further processing,
\lidar{} point clouds require extensive processing to obtain object-level features.
These differences make processing radar point clouds different and sometimes more 
challenging compared to \lidar{} point clouds.

Some existing point-based proposal generation methods process point cloud by 
first projecting it to different views or using voxels to represent it in a compact 
form. 2D or 3D convolutional networks are then used to extract features. Other 
methods extract features from the raw point clouds directly using networks such as 
PointNet \cite{qi2017pointnet++}. These methods are usually designed for dense 
LIDAR point clouds and do not perform equally well on sparse radar point clouds.
Additionally, unlike \lidar{} point clouds, radar point clouds do not provide a 
precise 3D image of the object, as an ideal radar reports only one 
point for an object. Aggregating multiple radar readings obtained in different time-stamps
can help provide more points in the point cloud, but these points are not a good representation of
the objects' shape and size. Fig. \ref{fig:NuscSample} visualizes some of these differences
by showing radar and \lidar{} point clouds for a sample scene from the nuScenes dataset.

In this work, we propose a radar-camera fusion algorithm for joint object detection
and distance estimation in autonomous driving applications. The proposed method 
is designed as a two-stage object detection network that fuses radar point clouds
and learned image features to generate accurate object proposals. For every object 
proposal, a depth value is also calculated to estimate the object's distance from the 
vehicle. These proposals are then fed into the second stage of the detection network 
for object classification. We evaluate our network on the nuScenes dataset 
\cite{caesar2019nuscenes}, which provides synchronized data from multiple radar 
and camera sensors on a vehicle. Our experiments show that the proposed method 
outperforms other radar-camera fusion methods in the object detection task and 
is capable of accurately estimating distance for all detected objects.

\section{Related work} \label{relwork}
In this section we highlight some of the existing works on object detection and 
sensor fusion for autonomous vehicles, categorizing them into single-modality and 
fusion-based approaches.

\subsection{Single-Modality Object Detection}
Most vision-based object detection networks follow one of the two approaches:
two-stage or single-stage detection pipelines \cite{feng2019deep}. In two-stage 
detection networks, a set of class-agnostic object proposals are generated in the
first stage, and are refined, classified and scored in the second stage. R-CNN 
\cite{girshick2014rich} is the pioneering work in this category, using proposal 
generation algorithms such as Selective Search \cite{uijlings2013selective} in the 
first stage and a CNN-based detector in the second stage. Fast R-CNN 
\cite{girshick2015fast} also uses an external proposal generator, but eliminates 
redundant feature extraction by utilizing the global features extracted from the 
entire image to classify each proposal in the second stage.
Faster R-CNN \cite{ren2015faster} unifies the proposal generation and 
classification by introducing the Region Proposal Network (RPN), which uses the 
global features extracted from the image to generate object proposals.

One-stage object detection networks on the other hand directly map the extracted 
features to bounding boxes by treating the object detection task as a regression 
problem. YOLO \cite{redmon2016you} and SSD \cite{liu2016ssd} detection networks
are in this category, regressing bounding boxes directly from the extracted 
feature maps. One-stage detection networks are usually faster, but less 
accurate than their two-stage counterparts. By addressing the foreground-background 
class imbalance problem in single-stage object detection, RetinaNet \cite{lin2017focal}
achieved better results than the state-of-the-art two-stage detection networks.

Most of the point-based object detection networks focus on dense point clouds obtained 
from \lidar{}s. Some of these methods process the points by discretizing the 3D 
space into 3D voxels \cite{li20173d, sindagi2019mvx}, while others process the 
point clouds in the continuous vector space without voxelization to obtain individual 
features for each point \cite{qi2017pointnet,qi2017pointnet++}.
For object detection and classification using radar data, \cite{werber2015automotive}
proposes radar grid maps by accumulating radar data over several time-stamps, while
\cite{visentin2017classification} uses CNNs on a post-processed range-velocity map.
The radar data can also be processed as a 3D point cloud. \cite{danzer20192d} and 
\cite{schumann2018semantic} both use PointNet to perform 2D object classification
and segmentation, respectively.

\subsection{Fusion-based Object Detection}

Most fusion-based methods combine the \lidar{} point clouds with RGB images for 
2D or 3D object detection \cite{xu2018pointfusion, qi2018frustum}. In 
\cite{mv3d_2017} the network uses a multi-view 
representation of the 3D \lidar{} point clouds. The network projects the points to the 
Bird's Eye View (BEV) and front view planes, and uses the BEV to generate object 
proposals. \cite{ji2008radar} projects radar detections to the image and generate 
object proposals for a small CNN classification network. In \cite{nabati2019rrpn}, authors 
map radar detection to the image plane and use a radar-based RPN to generate 2D 
object proposals for different object categories in a two-stage object detection 
network. Authors in \cite{nobis2019deep} also project radar detections to the 
image plane, but represent radar detection characteristics as pixel values. 
The RGB image is then augmented with these values and processed in a CNN
to regress 2D bounding box coordinates and classification scores.

\begin{figure*}[th]
   \centering
   \includegraphics[width=0.99\textwidth]{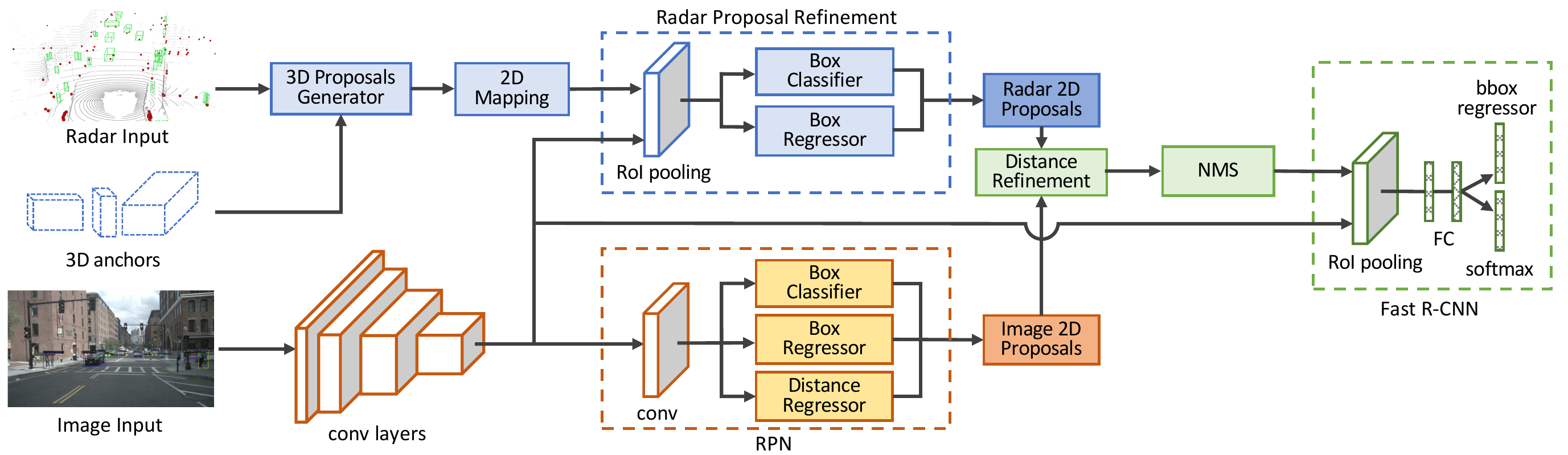}
   \caption{The proposed network architecture. Inputs to the network are radar point 
      cloud, camera image and 3D anchor boxes. radar-based object proposals are 
      generated from the point cloud and fused with image features to improve box 
      localization.}
   \label{fig: Arch}
\end{figure*}

\section{Our Framework} \label{framework}
Our proposed sensor fusion network is shown in Fig. \ref{fig: Arch}. The network takes
radar point clouds and RGB images as input and generates accurate object proposals
for a two-stage object detection framework. We take a middle-fusion approach for 
fusing the radar and image data, where outputs of each sensor are processed independently 
first, and are merged at a later stage for more processing. More specifically, we first use the
radar detections to generate 3D object proposals, then map the proposals to the image 
and use the image features extracted by a backbone network to improve their localization.
These proposals are then merged with image-based proposals generated in a RPN, and 
are fed to the second stage for classification. All generated proposals are associated with an
estimated depth, calculated either directly from the radar detections, or via a distance regressor
layer in the RPN network.

\subsection{Radar Proposal Network}

Our proposed architecture treats every radar point as a stand-alone detection and 
generates 3D object proposals for them directly without any feature extraction. These
proposals are generated using predefined 3D anchors for every object class in 
the dataset. Each 3D anchor is parameterized as $(x,y,z,w,l,h,r)$,
where $(x,y,z)$ is the center, $(w,l,h)$ is the size, and $(r)$ is the orientation of 
the box in vehicle's coordinate system. The anchor size, $(w,l,h)$, is 
fixed for each object category, and is set to the average size of the objects in 
each category in the training dataset. 
For every anchor box, we use two different orientations, r = \{0\degree, 90\degree\} 
from the vehicle's centerline. The center location for each anchor is obtained from the 
radar detection's position in the vehicle coordinates. For every radar point, we generate \textit{2n} 
boxes from the 3D anchors, where \textit{n} is the number of object classes in the 
dataset, each having two different orientations.

In the next step, all 3D anchors are mapped to the image plane and converted 
to equivalent 2D bounding boxes by finding the smallest enclosing box for 
each mapped anchor. Since every 3D proposal is generated from a radar detection, 
it has an accurate distance associated with it. This distance is used as the proposed 
distance for the generated 2D bounding box. Since 3D anchors with the same size as 
objects of interest are used to generate 
the 2D object proposals on the image, the resulting proposals capture 
the true size of the objects as they appear in the image. This eliminates the need 
for adjusting the size of radar proposals based on their distance from the vehicle, 
which was proposed in \cite{nabati2019rrpn}.

\begin{figure*}[th!]
   \centering
   \subfloat[]{
      \includegraphics[width=0.24\textwidth]{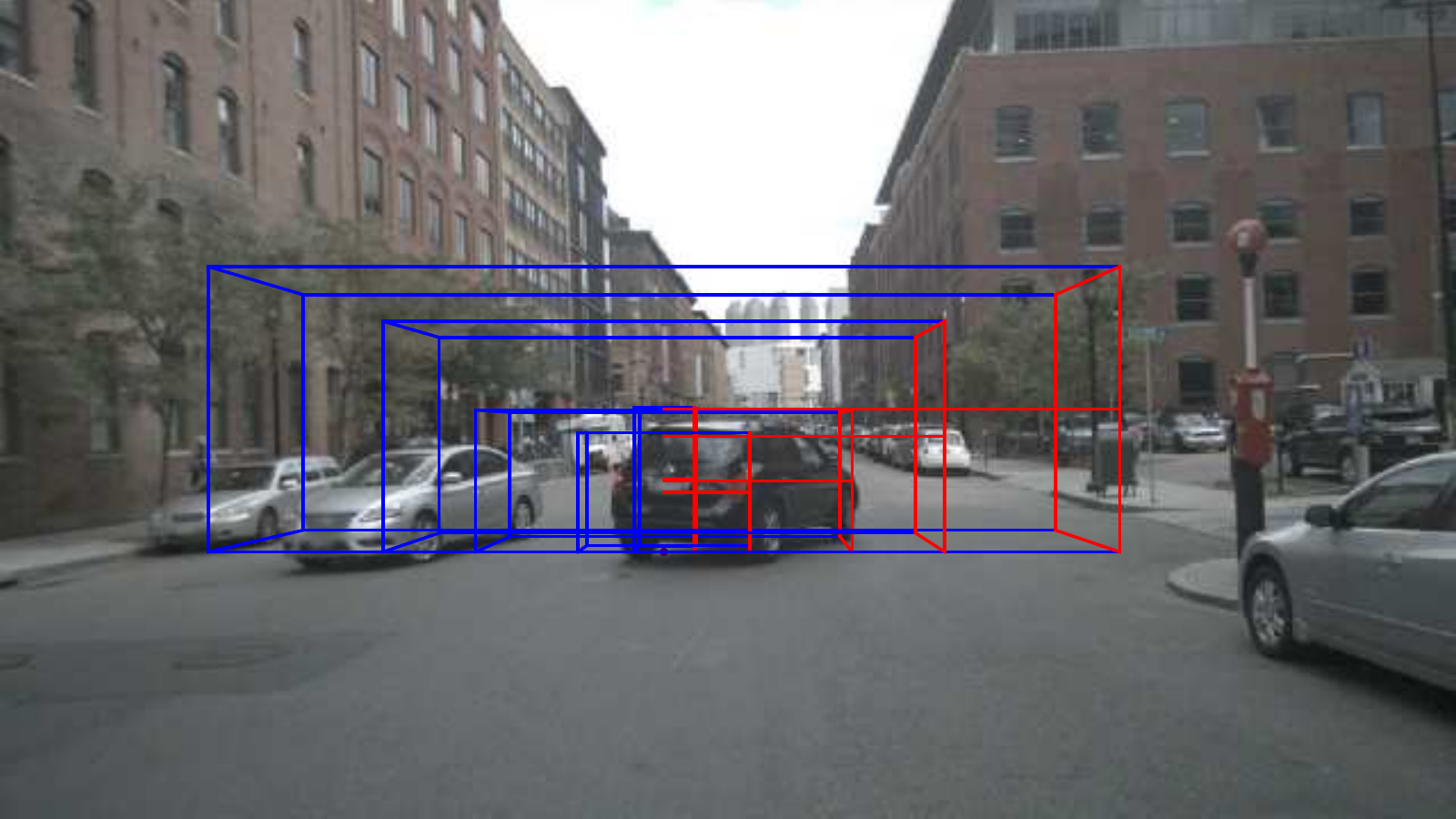}}
      \label{1}
   \hspace{-1mm}%
   \subfloat[]{
      \includegraphics[width=0.24\textwidth]{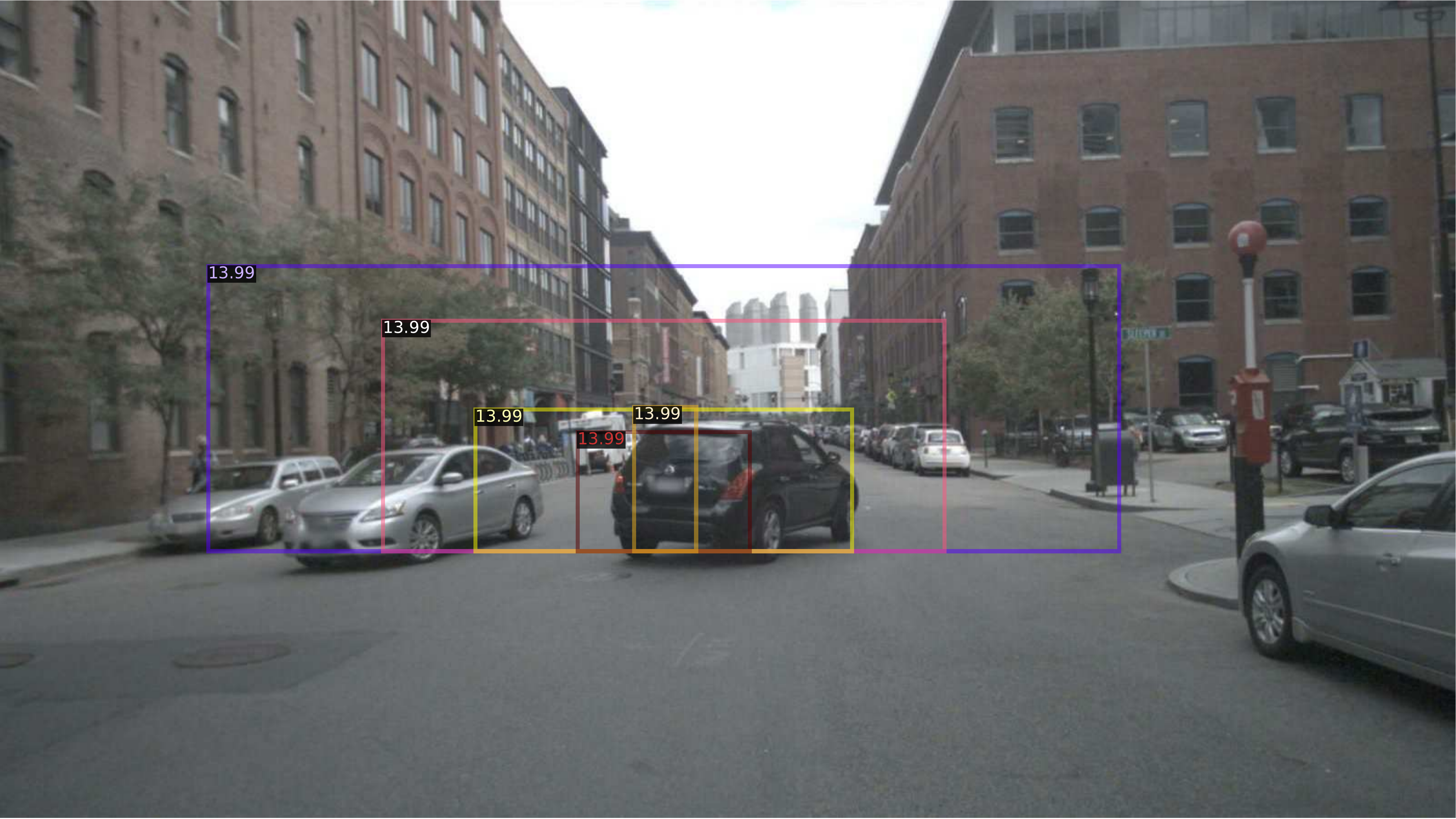}}
      \label{2}
   \hspace{-1mm}%
   \subfloat[]{
         \includegraphics[width=0.24\textwidth]{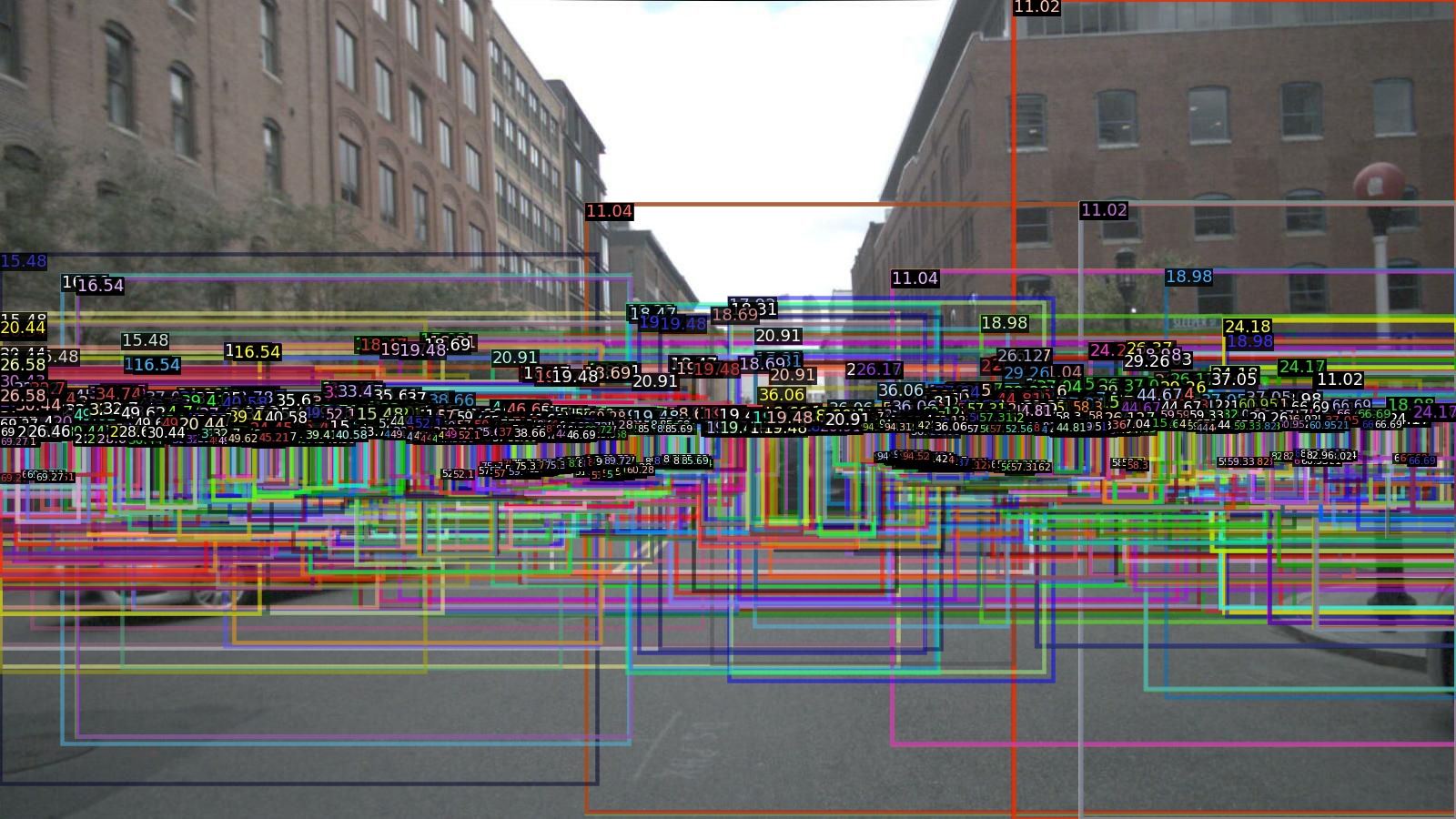}}
         \label{3}
   \hspace{-1mm}%
   \subfloat[]{
         \includegraphics[width=0.24\textwidth]{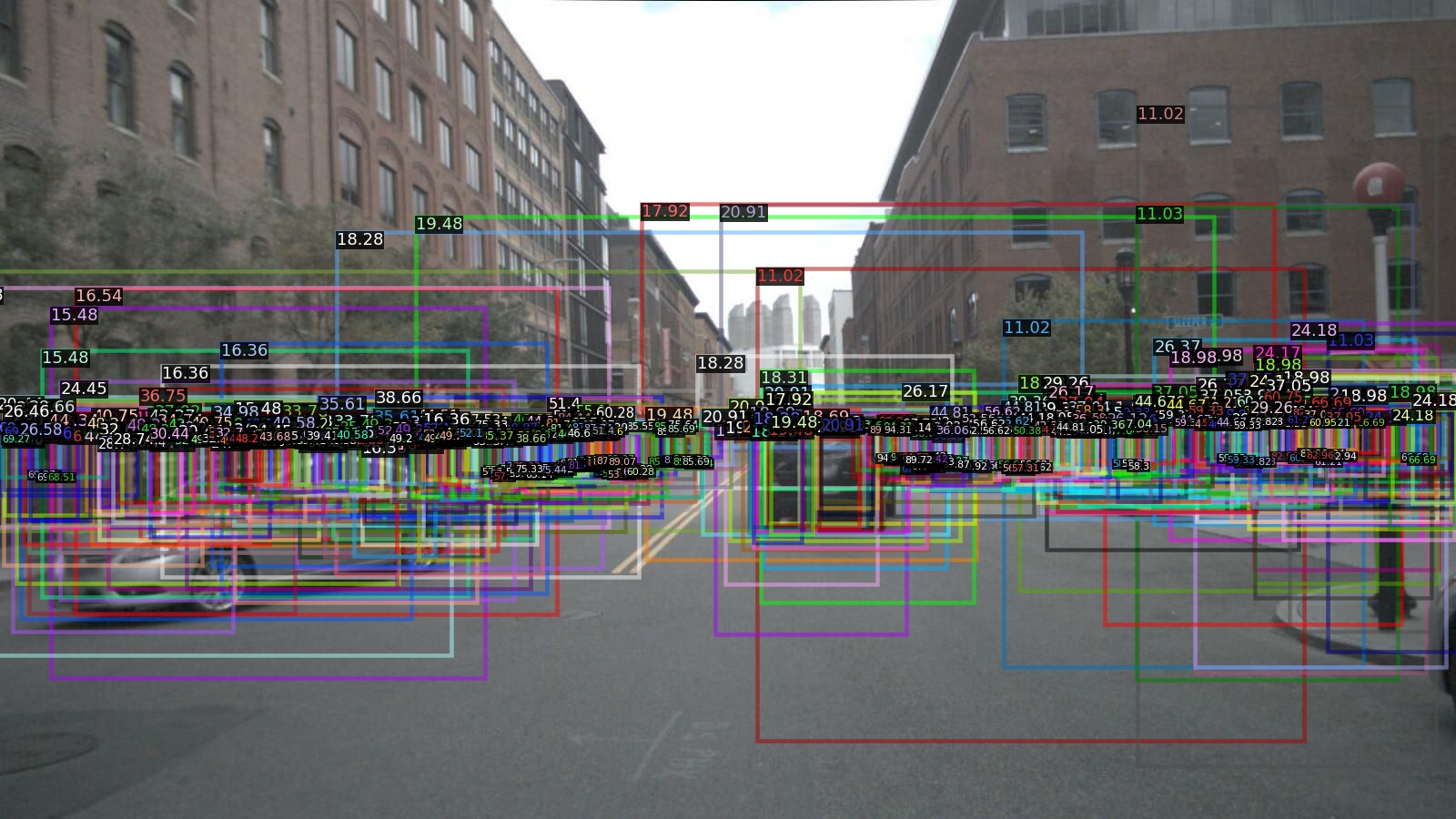}}
         \label{4}
   \caption{Radar-based proposals. 
      (a): 3D anchors for one radar detection ($r=90\degree$). 
      (b): 2D proposals obtained from 3D anchors.
      (c): 2D proposals for all radar detections inside the image. 
      (d): Refined radar proposals after applying box regression.
      Radar-based distances in meters are shown on the bounding boxes.}
   \label{fig:propBox}
\end{figure*}

Fig. \ref{fig:propBox}(b) illustrates 3D anchors and equivalent 2D proposals 
generated for a sample image. As shown in this figure, radar-based proposals are 
always focused on objects that are on the road plane. This prevents unnecessary 
processing of areas of the image where no physical object exists, such as the sky 
or buildings in this image.

In the next step, all generated 2D proposals are fed into the Radar Proposal 
Refinement (RPR) subnetwork. This is where the information obtained from the 
radars (radar proposals) is fused with the information obtained from the 
camera (image features). RPR uses the features extracted from the image by the 
backbone network to adjust the size and location of the radar proposals on the image.
As radar detections are not always centered on the corresponding objects on the image,
the generated 3D anchors and corresponding 2D proposals might be offset as well.
The box regressor
layer in the RPR uses the image features inside each radar proposal to regress 
offset values for the proposal corner points. The RPR also contains a box 
classification layer, which estimates an objectness 
score for every radar proposal. The objectness score is used to eliminate proposals 
that are generated by radar detections coming from background objects, such as 
buildings and light poles. The inputs to the box regressor and classifier layers are 
image features inside negative and positive radar proposals. We follow \cite{ren2015faster}
and define positive proposals as ones with an Intersection-over-Union (IoU) overlap 
higher than 0.7 with any ground truth bounding box, and negative proposals as 
ones with an IoU below 0.3 for all ground truth boxes. Radar proposals with an 
IoU between 0.3 and 0.7 are not used for training. Since radar proposals have different 
sizes depending on their distance, object category and orientation, a RoI Pooling 
layer is used before the box regression and classification layers to obtain feature 
vectors of the same size for all proposals. Fig. \ref{fig:propBox}(d) 
shows the radar proposals after the refinement step.

\subsection{Image Proposal Network}
Our architecture also uses a RPN network to generate object proposals from the 
image. The radar proposal network is not always successful in generating proposals for 
certain object categories that are harder for radars to detect but are easily 
detected in the image, such as pedestrian or bicycles. On the other hand, 
the image-based proposal network might fail 
to detect far away objects that are easily detected by the radar.
Having an image-based object proposal network in addition to the radar-based 
network improves the object detection accuracy, as they complement each other by 
using two different modalities for proposal generation and distance estimation. 

Image-based object proposals are generated by a network similar to the RPN 
introduced in Faster R-CNN \cite{ren2015faster}. The input to this network is the 
image feature maps extracted by the backbone CNN. To estimate 
distance for every object proposal, we add a fully connected distance regression 
layer on top of the convolutional layer in RPN, as shown in Fig. \ref{fig: Arch}.
This layer is implemented with a 1$\times$1 convolutional layer similar to the 
box-regression and box-classification layers in the RPR network. However, 
because it's difficult to directly regress to distance from an image, we use the 
output transformation of Eigen \textit{et. al} \cite{eigen2014depth} and use 
$d = \frac{1}{\sigma(\hat{d})} -1$ where $\hat{d}$ is 
the regressed distance value. The distance regression layer generates \textit{k} 
outputs, where \textit{k} is the number of 2D anchor boxes used in the RPN network
at each location on the feature map. We use a cross entropy loss for object 
classification and a Smooth L1 loss for box distance regressor layers.

\subsection{Distance Refinement}
The outputs of the radar and image proposal networks need to be merged for the 
second stage of the object detection network. Before using the proposals in the 
next stage, redundant proposals are removed by applying Non-Maximum Suppression 
(NMS). The NMS would normally remove overlapping proposals without discriminating 
based on the bounding box's origin, but we note that radar-based proposals have more 
reliable distance information than the image-based proposals. This is because 
image-based distances are estimated only from 2D image feature maps with no 
depth information. To make sure the radar-based distances are not unnecessarily 
discarded in the NMS process, we first calculate the Intersection over Union (IoU) between 
radar and image proposals. Next we use an IoU threshold to find the matching proposals, 
and overwrite the image-based distances by their radar-based 
counterparts for these matching proposals. The calculated IoU values are reused
in the next step where NMS is applied to all proposals, regardless of their origin.
The remaining proposals are then fed into the second stage of the detection network 
to calculate the object class and score.

\subsection{Second Stage Detection Network}
The inputs to the second stage detection network are the feature map from the
image and object proposals. The structure of this network is similar to 
Fast R-CNN \cite{girshick2015fast}. The feature map is cropped for every object 
proposals and is fed into the RoI pooling layer to obtain feature vectors of the 
same size for all proposals. These feature vectors are further processed by a 
set of fully connected layers and are passed to the softmax and bounding box 
regression layers. The output is the category classification and bounding box 
regression for each proposal, in addition to the distance associated to every detected object. 
Similar to the RPN network, we use a cross entropy
loss for object classification and a Smooth L1 loss for the box regression layer.

\subsection{Loss Function}
We follow Faster R-CNN \cite{ren2015faster} and use the following multi-task loss 
as our objective function:

\begin{equation*}
   L({p_i},{t_i}) = \dfrac{1}{N_{cls}} \sum_{i}L_{cls}(p_i, p^*_i) + \ 
   \lambda \dfrac{1}{N_{reg}} \sum_{i}p^*_i L_{reg}(t_i, t^*_i).
\end{equation*}

where $i$ is the anchor index, $p_i$ is the $i$'th anchor's objectness score,
$p^*_i$ is the ground truth score (1 if anchor is positive and 0 if negative),
$t_i$ is the vector of 4 parameters representing the predicted bounding box and 
$t^*_i$ is the ground truth bounding box. We use the log loss over two classes 
for the classification loss $L_{cls}$, and the the smooth $L_1$ loss for the 
regression loss, $L_{reg}$. $N_{cls}$ and $N_{reg}$ are normalization factors and 
$\lambda$ is a balancing parameter.

\section{Experiments}

\subsection{Dataset and Implementation Details}
Our network uses FPN \cite{lin2017focal} with ResNet-50 \cite{he2016deep} pretrained 
on ImageNet as the backbone for image feature extraction. We use the same RPN 
architecture as Faster R-CNN \cite{ren2015faster}, and only add the distance regression layer 
on top of its convolution layer for distance estimation. For the second stage of the network, the 
classification stage, we use the same architecture as Fast R-CNN.

We use the nuScenes dataset \cite{caesar2019nuscenes} to evaluate our network. 
Out of 23 different object classes in this dataset, we use 6 classes as shown in 
Table \ref{table:results2}. The nuScenes dataset includes data from 6 different cameras 
and 5 radars mounted on the vehicle. We use samples from the front- and rear-view 
cameras together with detection from all the radars for both training and evaluation. 
The ground truth annotations in the nuScenes dataset are 
provided in the form of 3D boxes in the global coordinate system. As a 
preprocessing step, we first transform the annotations and radar point clouds to the 
vehicle coordinate, then convert all 3D 
annotations to their equivalent 2D bounding boxes.
This is achieved by mapping the 3D boxes to the image and finding the 
smallest 2D enclosing bounding box. For every 3D annotation, we also calculate 
the distance from vehicle to the box and use it as the ground truth distance for its
2D counterpart. The official nuScenes splits are used for training and evaluation,
and images are used at their original resolution (900$\times$1600) for both steps. No data 
augmentation is used as the number of labeled instances for each category is 
relatively large. We used PyTorch to implement our network and all 
experiments were conducted on a computer with two Nvidia Quadro P6000 GPUs.

\begin{table}[t]
   \scriptsize
   \centering
   \caption{Performance on the nuScenes validation set. }
   \setlength\tabcolsep{3pt}
   \begin{tabular}{l|c|c|c|c|c|c}
         {}   & Weighted AP & AP & AP50 & AP75 & AR & MAE \\
      \hline\hline
      Faster R-CNN & No & 34.95 & 58.23 & 36.89 & 40.21 & - \\
      RRPN     & No  & 35.45 & 59.00 & 37.00 & \textbf{42.10} &  - \\
      Ours     & No  & \textbf{35.60}&\textbf{60.53} &\textbf{37.38}&\textbf{42.10}& 2.65 \\
      \hline
      Faster R-CNN & Yes & 43.78 &   -   &  -  &  -  &  - \\
      CRF-Net  & Yes & 43.95 &   -   &  -  &  -  &  - \\
      Ours     & Yes & \textbf{44.49} &   -   &  -  &  -  &  -  \\
   \end{tabular}
   \footnotetext{Text in the footnote}
   \label{table:results1}
\end{table}

\begin{table}[t]
   \scriptsize
   \centering
   \caption{Per-class performance}
   \setlength\tabcolsep{3pt}
   \begin{tabular}{c|c|c|c|c|c|c}
       {} & Car & Truck & Person & Bus & Bicycle & Motorcycle \\
      \hline\hline
      Faster R-CNN  & 51.46 & 33.26 & 27.06 & 47.73 & 24.27 & 25.93\\ %
      RRPN     & 41.80 & \textbf{44.70} & 17.10 & \textbf{57.20} & 21.40 & \textbf{30.50}\\ %
      Ours     & \textbf{52.31} & 34.45 & \textbf{27.59} & 48.30 & \textbf{25.00} & 25.97\\ %

   \end{tabular}
   \footnotetext{Text in the footnote}
   \label{table:results2}
\end{table}

\begin{table}[t!]
   \scriptsize
   \centering
   \caption{Per-class Mean Absolute Error (MAE) for distance estimation}
   \setlength\tabcolsep{3pt}
   \begin{tabular}{l|c|c|c|c|c|c}
      Category & Car & Truck & Person & Bus & Bicycle & Motorcycle \\
      \hline
      MAE & 2.66 & 3.26 & 2.99 & 3.187 & 1.97 & 2.81\\
   \end{tabular}
   \label{table:MAE}
\end{table}

\begin{figure*}[th]
   \subfloat{\includegraphics[width=0.33\textwidth]{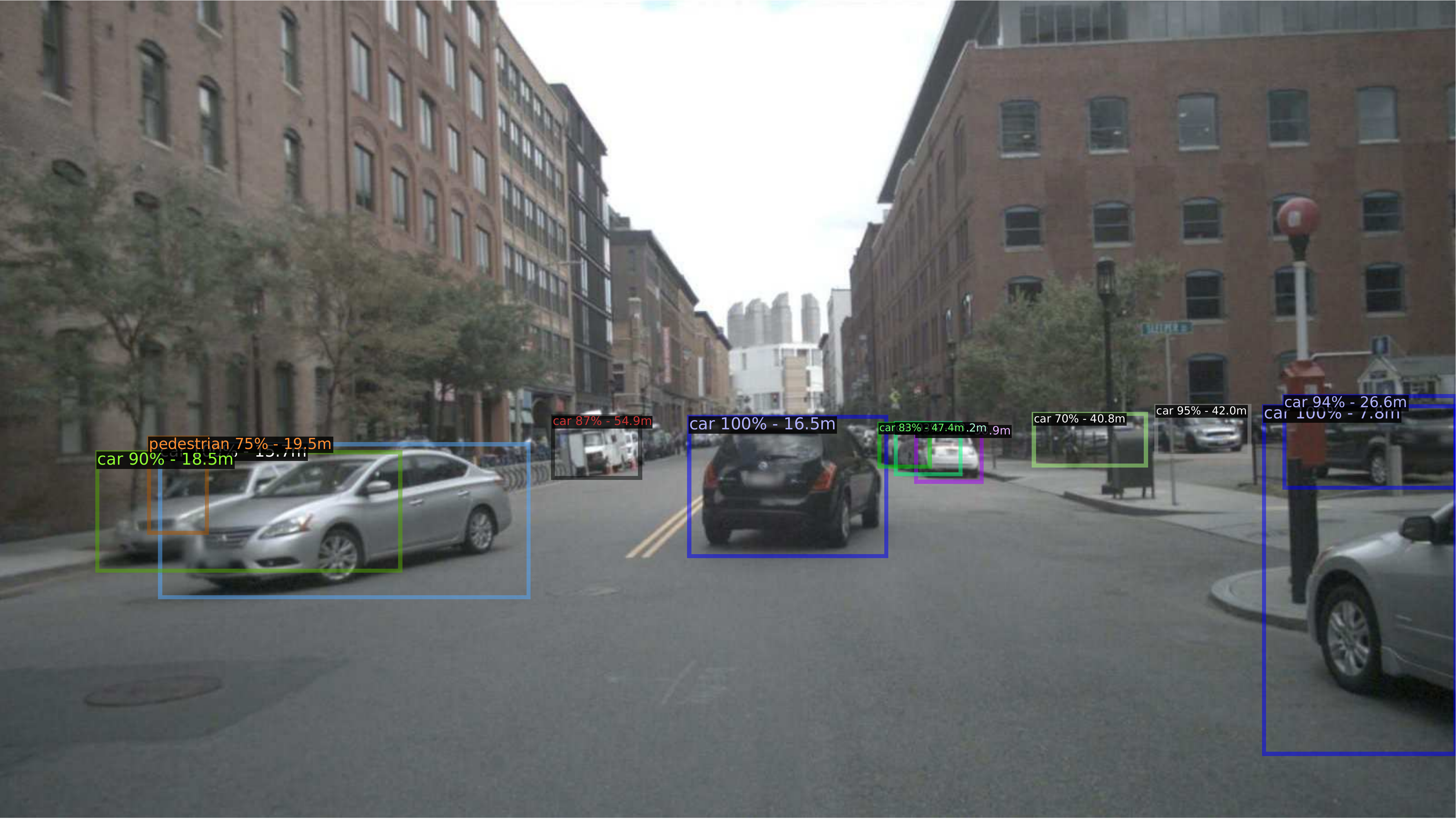}}\hfill
   \subfloat{\includegraphics[width=0.33\textwidth]{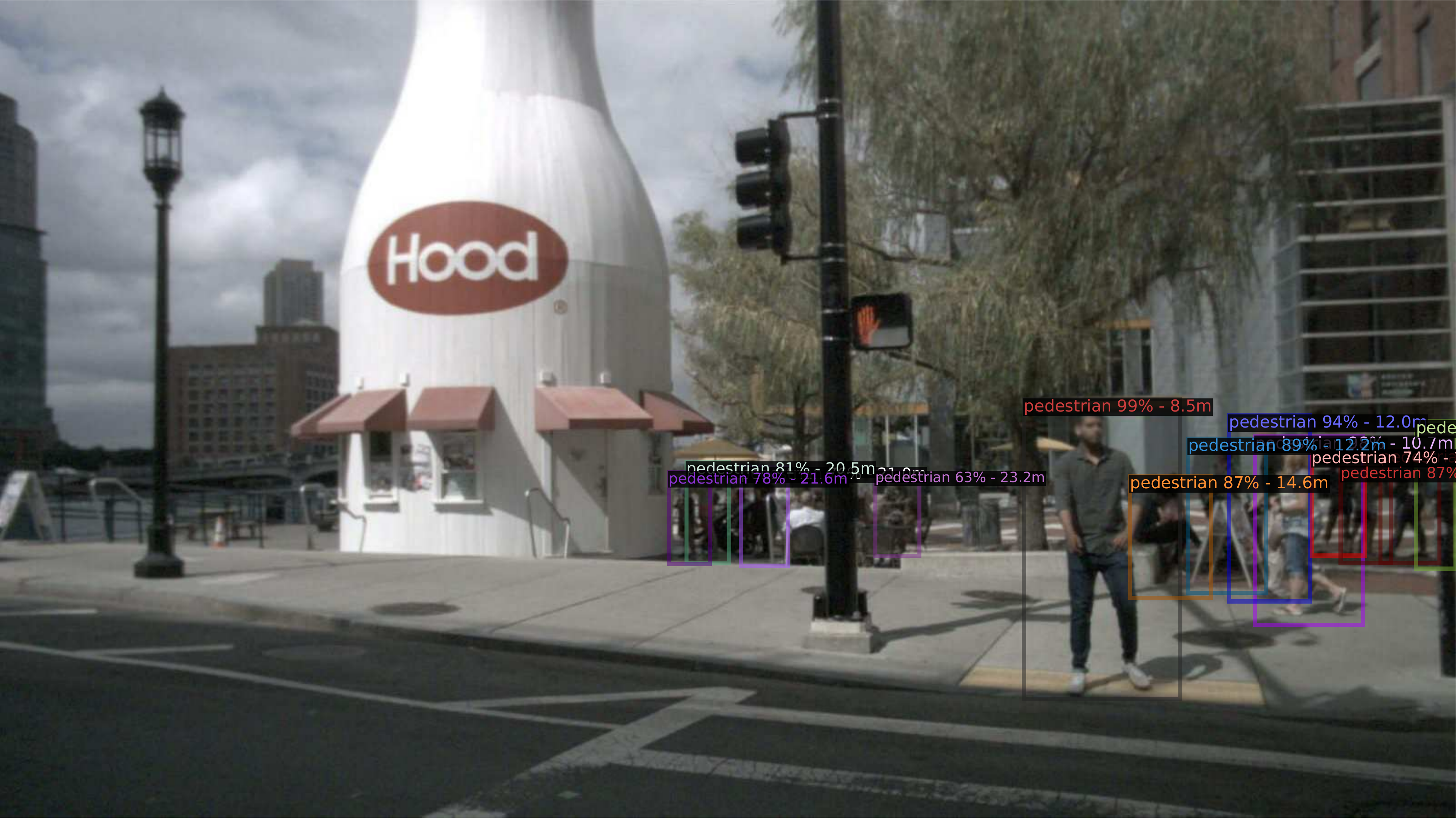}}\hfill
   \subfloat{\includegraphics[width=0.33\textwidth]{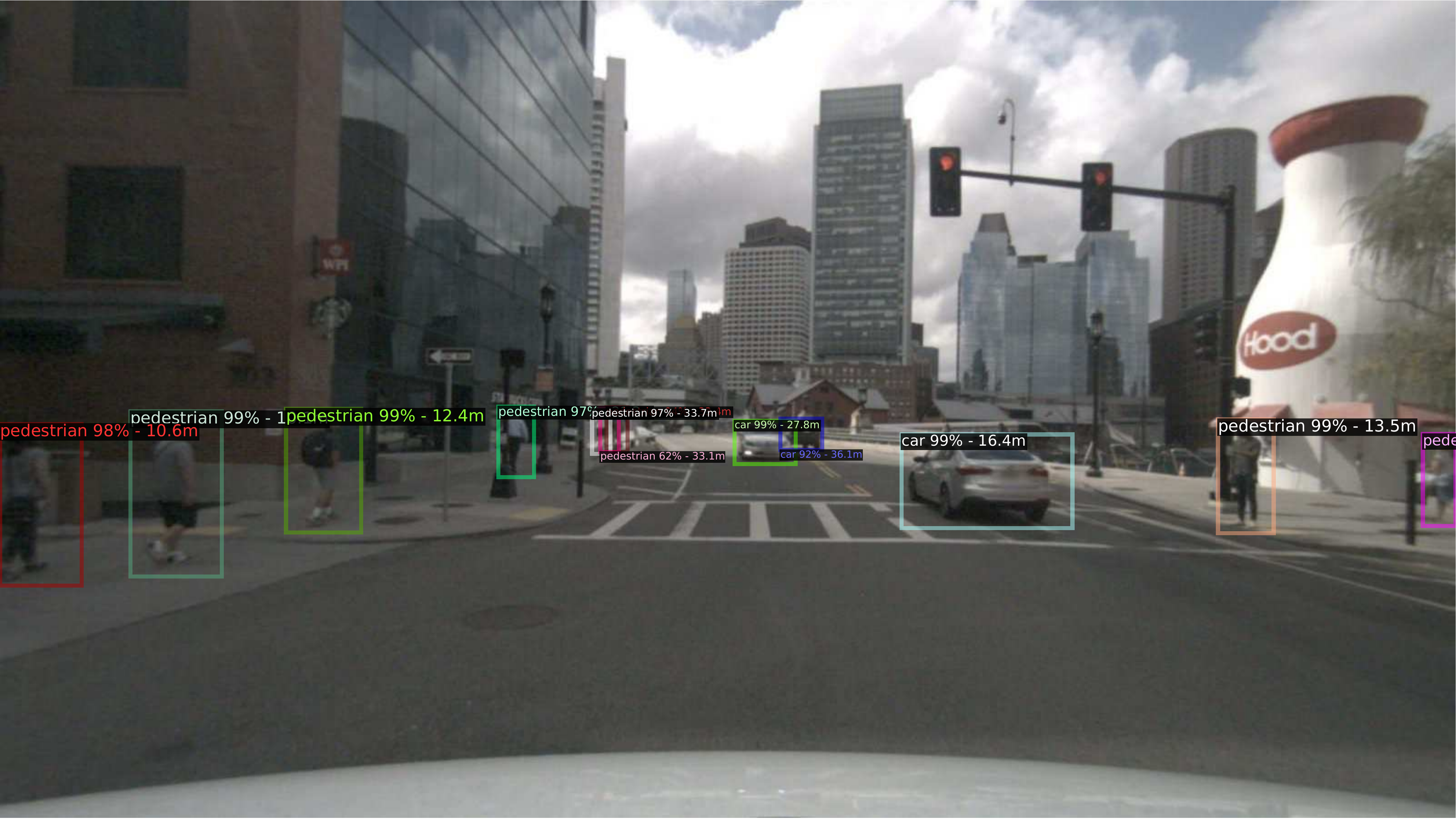}}\\[-1.5ex]
   \subfloat{\includegraphics[width=0.33\textwidth]{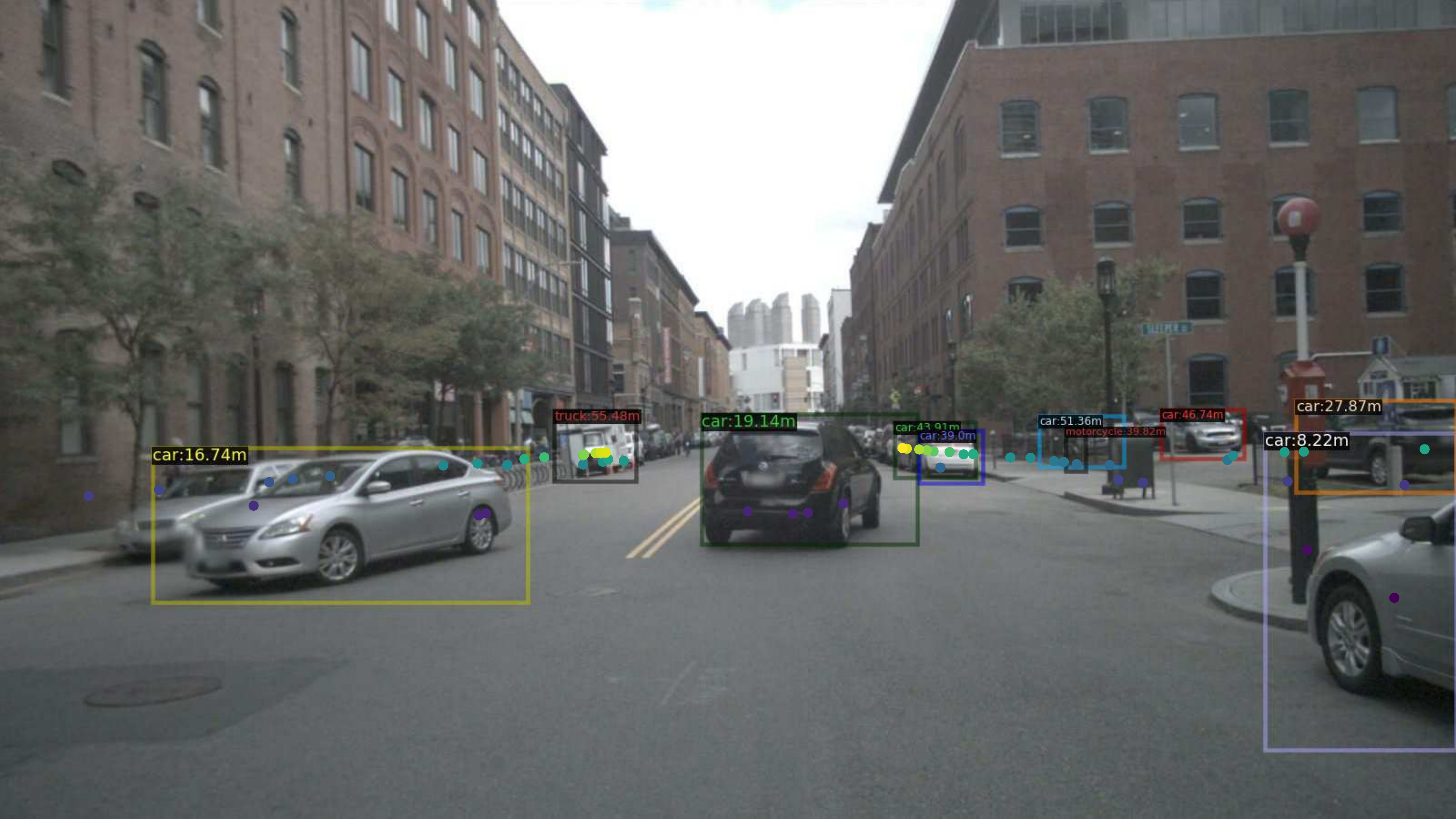}}\hfill
   \subfloat{\includegraphics[width=0.33\textwidth]{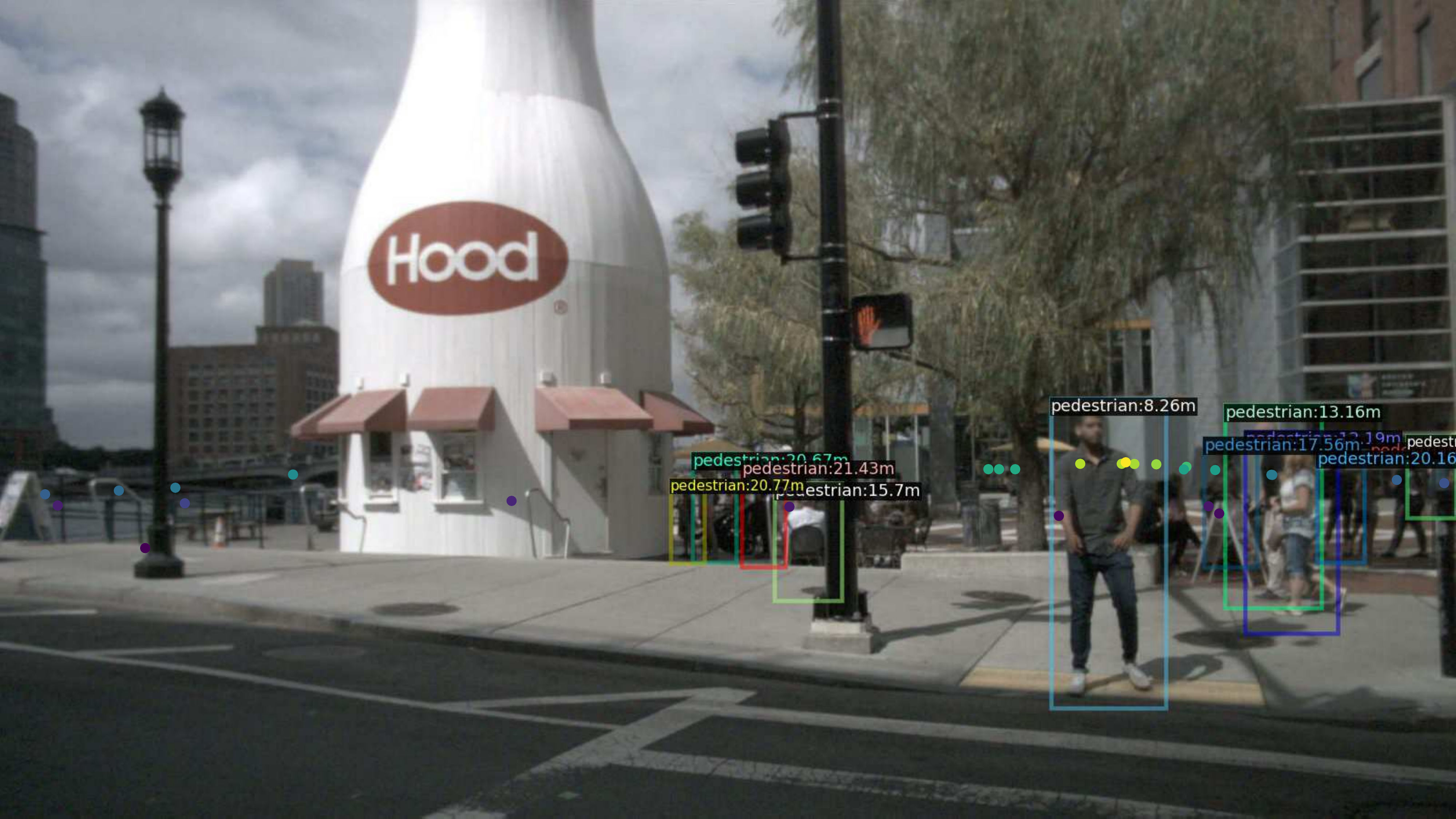}}\hfill
   \subfloat{\includegraphics[width=0.33\textwidth]{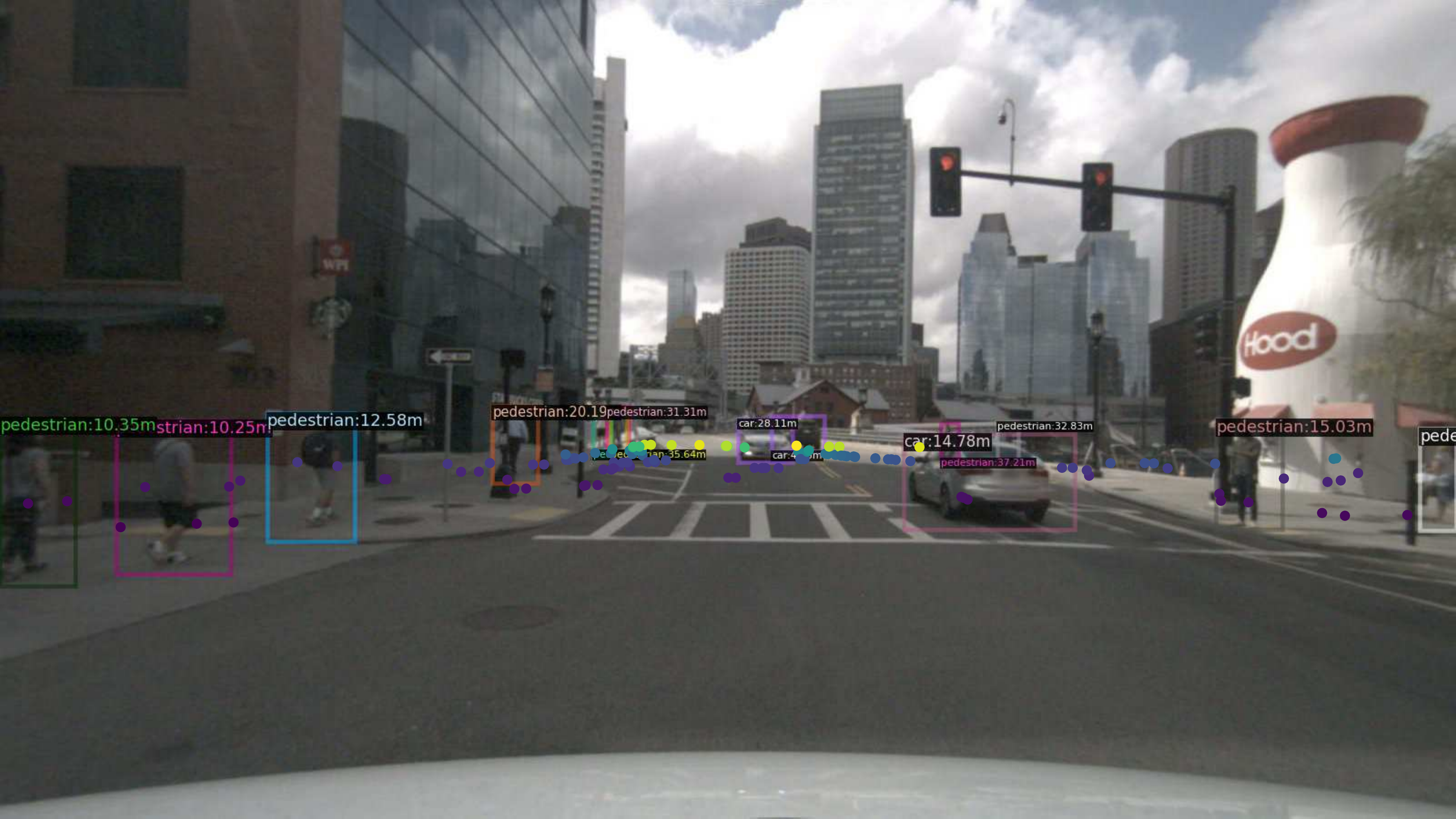}}
   \caption{Object detection and distance estimation results. Top: detection 
      outputs, Bottom: ground truth. (Best viewed in color and zoomed-in)}
   \label{results}
\end{figure*}

\subsection{Evaluation}
The performance of our method is shown in Table \ref{table:results1}. This table 
shows the overall Average Precision (AP) and Average Recall (AR) for the detection task,
and Mean Absolute Error for the distance estimation task. We use the Faster R-CNN 
network as our image-based detection baseline, and compare our results with 
RRPN \cite{nabati2019rrpn} and CRF-Net\cite{nobis2019deep}, which use 
radar and camera fusion for object detection.
CRF-Net only uses images from the front-view camera and also uses a weighted AP 
score based on the number of object appearances in the dataset. For fair comparison, 
we use the weighted AP scores to compare our results with this network.
The CRF-Net also reports some results 
after filtering the ground truth to consider only objects that are detected by 
at least one radar, and filtering radar detections that are outside 3D 
ground truth bounding boxes. We do not apply these filtering operations and only 
compare with their results on the unfiltered data. Since CRF-Net does not 
report AR, per-class AP, or AP for different IoU levels, we only compare our overall
AP with theirs.

According to Table \ref{table:results1} our method outperforms RRPN and CRF-Net for 
the detection task, improving the AP score by 0.15 and 0.54 points respectively.
Our proposed method also accurately estimates the distance for all 
detected objects, as visualized in Fig. \ref{results}. We use Mean Absolute Error 
(MAE) as the evaluation metric for distance estimation. Our method achieves an MAE
of 2.65 on all images. The per-class MAE values are provided in Table \ref{table:MAE}. 
According to this table, larger objects such as trucks and buses have a higher 
distance error compared to other classes. This behavior 
is expected and could be explained by the fact that radars usually report 
multiple detections for larger objects, which results
in several object proposals with different distances for the same object. Additionally, 
most radar detections happen to be at the edge of objects, while the ground truth 
distances are measured from the center of objects. This results in higher distance 
mismatch error for larger objects, where the distance between the edge and center of 
the object is significant.

\section{Conclusion and Future Work}
We proposed a radar-camera fusion algorithm for joint object detection and 
distance estimation for autonomous driving scenarios. The proposed architecture 
uses a multi-modal fusion approach to employ radar point clouds and image 
feature maps to generating accurate object proposals. The proposed network also 
uses both radar detections and image features for distance estimation for 
every generated proposal. These proposals are fed into the second stage of the 
detection network for object classification.
Experiments on the nuScenes dataset show that our method outperforms other 
radar-camera fusion-based object detection methods, while at the same time accurately
estimates the distance to every detection.

As a future work, we intend to work on reducing the distance error introduced by the mismatch 
between radar detections and ground truth measurements. 
This can be alleviated to some extent by a pre-processing step, where the ground truth 
distances are re-calculated based on the distance between the edge of the bounding 
boxes to the vehicle. Additionally, a clustering algorithm could be used to group 
the Radar detections and reduce the distance error introduced by having multiple 
detections for larger objects.

\bibliographystyle{IEEEtran}
\bibliography{ref}

\end{document}